\documentclass{article}
\usepackage{cite}
\usepackage{spconf,amsmath,epsfig}
\usepackage{subcaption}
\let\OLDthebibliography\thebibliography
\renewcommand\thebibliography[1]{
  \OLDthebibliography{#1}
  \setlength{\parskip}{0pt}
  \setlength{\itemsep}{0pt plus 0.3ex}
}

\begin{document}\sloppy

\def\x{{\mathbf x}}
\def\L{{\cal L}}

\title{Conv-INR: Convolutional Implicit Neural Representation for Multimodal Visual Signals}
%
\name{Zhicheng Cai}
%
%
\address{School of Electronic Science and Engineering, Nanjing University\\
\texttt{caizc@smail.nju.edu.cn}
}


\maketitle

\begin{abstract}
Implicit neural representation (INR) has recently emerged as a promising paradigm for signal representations. 
Typically, INR is parameterized by a multiplayer perceptron (MLP) which takes the coordinates as the inputs and generates corresponding attributes of a signal. However, MLP-based INRs face two critical issues: i) individually considering each coordinate while ignoring the connections; ii) suffering from the spectral bias thus failing to learn high-frequency components. While target visual signals usually exhibit strong local structures and neighborhood dependencies, and high-frequency components are significant in these signals, the issues harm the representational capacity of INRs.
This paper proposes Conv-INR, the first INR model fully based on convolution. 
Due to the inherent attributes of convolution, Conv-INR can simultaneously consider adjacent coordinates and learn high-frequency components effectively. Compared to existing MLP-based INRs, Conv-INR has better representational capacity and trainability without requiring primary function expansion. We conduct extensive experiments on four tasks, including image fitting, CT/MRI reconstruction, and novel view synthesis, Conv-INR all significantly surpasses existing MLP-based INRs, validating the effectiveness. Finally, we raise three reparameterization methods that can further enhance the performance of the vanilla Conv-INR without introducing any extra inference cost.
\end{abstract}

\begin{figure*}[htbp]
  \centering
  \includegraphics[width=.9\textwidth]{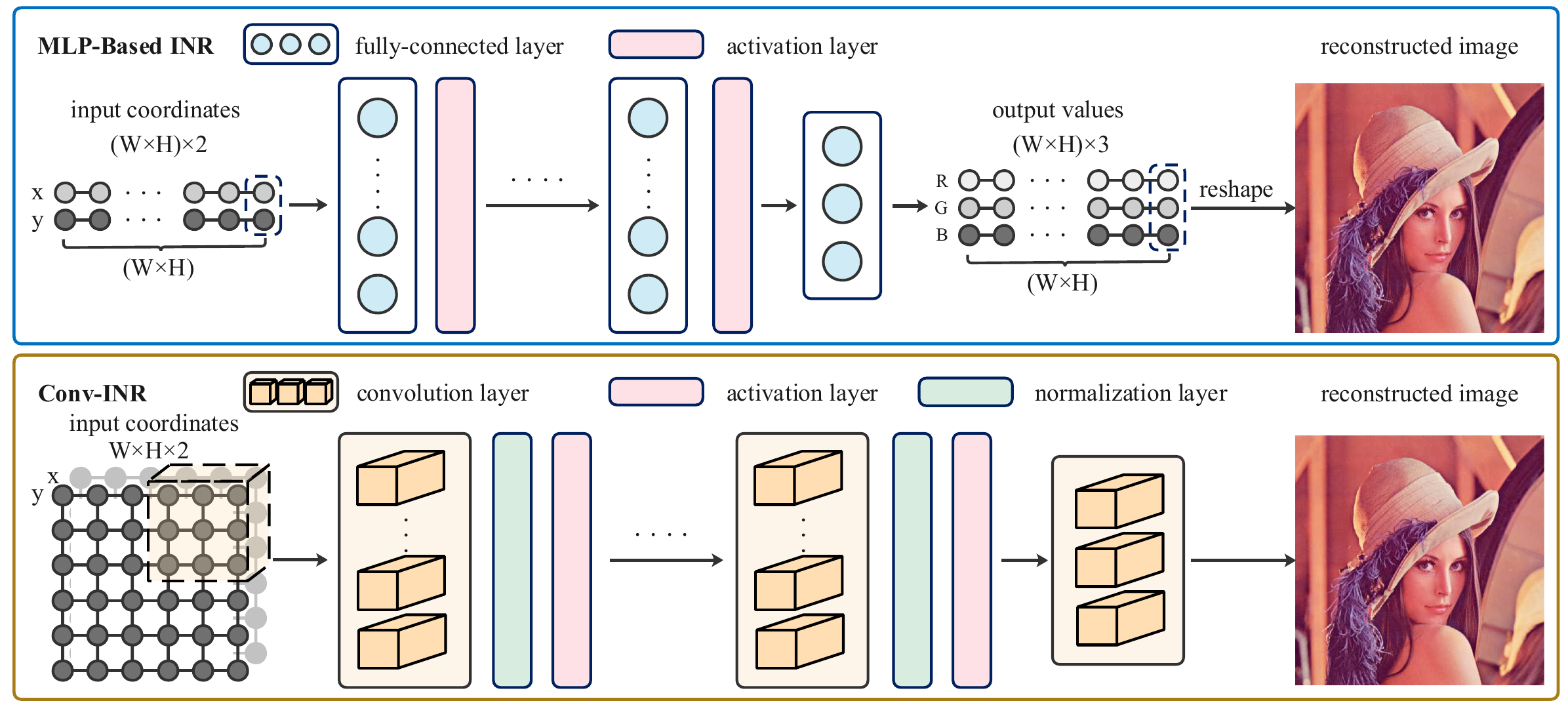} 
  \vspace{-1em}
  \caption{Pipelines of MLP-based INR and Conv-INR. Take the image fitting task as the example.}
  \label{p0}
  \vspace{-1em} 
\end{figure*}

\vspace{-1em}
\section{Introduction}
\vspace{-1em}
Implicit neural representation (INR), which characterizes a signal by preserving the mapping function from the coordinates to corresponding signal attributes using neural networks, has recently emerged as a promising signal processing framework and achieved great success in numerous vision-related tasks, including visual data representation and compression~\cite{chen2021nerv,niemeyer2020differentiable,park2019deepsdf, lindell2022bacon, dupont2021coin,dupont2022coin++,strumpler2022implicit}, scientific computing~\cite{xu2022signal,navon2023equivariant,karniadakis2021physics}, novel view synthesis~\cite{mildenhall2021nerf,pumarola2021d,barron2021mip}, and so on~\cite{ tancik2020fourier,sitzmann2020implicit,saragadam2023wire,fathony2020multiplicative}. 

Typically, INR is parameterized by a multiplayer perceptron (MLP) which takes each coordinate as the input and generates the corresponding value. 
However, MLP-based INRs suffer from the issue of spectral bias~\cite{basri2020frequency,rahaman2019spectral}, which makes MLP-based INRs converge faster to the low-frequency components while failing to effectively learn the high-frequency components of the signal.
To address this issue, two methods based on primary function expansion are proposed.
One is the positional encoding~\cite{mildenhall2021nerf, tancik2020fourier}, which aims at embedding multiple orthogonal Fourier or Wavelet bases~\cite{fathony2020multiplicative} into the subsequent network. However, a significant challenge arises from the fact that the frequency distribution of a signal to be inversely solved is often unknown. This can potentially lead to a mismatch between the pre-defined bases’ frequency set and the characteristics of the signal itself, resulting in an imperfect representation~\cite{yuce2022structured}.
The other is automatic frequency tuning, achieved through the use of periodic~\cite{sitzmann2020implicit} or non-periodic~\cite{saragadam2023wire} activation functions. Nevertheless, the supported frequency set is still limited and the representational capacity requires further improvement.

Moreover, considering the fact that the target visual signals, such as images, usually exhibit strong local structures and neighborhood dependencies, offering an important inductive bias for learning the representations of complex data~\cite{battaglia2018relational,lee2023locality,jiang2020local}, thus the adjacent spatial pixels are favorable to be considered together rather than in isolation~\cite{aftab2022multi}. 
However, existing MLP-based INRs separately take the pixels into computation, thus disregarding their relationships and failing to benefit from the locality prior, which hinders the representational capacity of the INRs to be further enhanced.

This paper raises Conv-INR, the first INR model fully based on convolution.
As shown in Fig~\ref{p0}, Conv-INR takes the $H\!\times\!W\!\times\!C$ coordinates tensor as the input and generates the represented signal directly. Due to the inherent sliding window mechanism of the convolution operator, pixels in different patches share the same convolution kernel, thus Conv-INR actually performs stationary (shift-invariant) convolution over the input domain and aggregates local information, it avoids suffering from the issue of spectral bias~\cite{tancik2020fourier,si2022inception,wang2020high,yin2019fourier}, making it capable of representing both low- and high-frequency components of the signal effectively without requiring complex primary function expansion.
Furthermore, unlike MLP-based INR solely focuses on an isolating pixel, Conv-INR simultaneously takes a patch of pixels into consideration, which effectively leverages the semantic prior of locality, thus obtaining better representational capacity. 

Extensive experiments are conducted on four tasks, i.e., image fitting, CT/MRI reconstruction and novel view synthesis, the vanilla Conv-INR all significantly surpasses existing MLP-based INRs (up to 6db PSNR higher than vanilla MLP INR on the Kodak image fitting dataset), including SIREN~\cite{sitzmann2020implicit}, PE-MLP~\cite{tancik2020fourier}, MFN~\cite{fathony2020multiplicative} and WIRE~\cite{saragadam2023wire}. 
The experimental results validate the SOTA representational capacity and generality of Conv-INR among existing INRs.

Nevertheless, we intend to further enhance the representational capacity of the vanilla Conv-INR. Thus for the first time, we introduce the \textit{reparameterization technique}~\cite{ding2019acnet,ding2021diverse,ding2021repvgg,ma2020weightnet,li2022omni,yang2019condconv,cai2023refconv,cao2022conv,arora2018optimization} into the realm of INR. We proposed three reparameterization methods specifically tailored for Conv-INR, including structural reparameterization~\cite{cai2023falconnet,ding2021diverse,ding2021repvgg}, static weight parameterization~\cite{cao2022conv,arora2018optimization} and dynamic weight reparameterization~\cite{li2022omni,yang2019condconv}, all bringing significant performance enhancement to the vanilla Conv-INR without introducing extra computational or memory cost for application.


\vspace{-1em}
\section{Background}
\vspace{-0.5em}
Firstly, let's review the mathematical formulation of INR.
Typically, INR is a MLP alternatively consisting of $L$ fully-connected layers $f^l$ and activation functions $\phi^l$, which can be expressed as:
\vspace{-0.7em}
\[
\vspace{-0.5em}
    \begin{split}
\vspace{-0.5em}
    F^{mlp}_{\theta}(\textbf{v}_n) = f^L\circ\phi^{L-1}\circ f^{L-1}\circ\cdot\cdot\cdot\circ\phi^{1}\circ f^{1}(\textbf{v}_n)
    \label{eq0}
\vspace{-0.5em}
    \end{split}
\vspace{-0.5em}
\]
Thus $F^{mlp}_{\theta}$ parameterized by $\theta$ learns the mapping from the normalized coordinate vector $\textbf{v}_n$ (e.g., $\textbf{v}_n\!=\!(x,y)$, $n\!=\!1,..., W\!\times\!H$ for image) to the corresponding intensity value $I(\textbf{v}_n)$. The weights $\theta$ are obtained by minimizing the following optimization problem:
\vspace{-0.7em}
\[
\vspace{-0.5em}
    \begin{split}
\vspace{-0.5em}
    \mathop{\arg\min}\limits_{\theta} \sum\limits_{\textbf{v}_n}\left|F^{mlp}_{\theta}(\textbf{v}_n)-I(\textbf{v}_n)\right|^2_2
    \label{eq1}
\vspace{-0.5em}
    \end{split}
\vspace{-0.5em}
\]
To overcome the spectral bias, two primary function expansion methods are raised. One is positional encoding~\cite{mildenhall2021nerf,tancik2020fourier} which encodes $\textbf{v}_n$ with $\gamma(\cdot)$ before passing into $F^{mlp}_{\theta}$:
\vspace{-0.7em}
\[
\vspace{-0.5em}
\begin{split}
\vspace{-0.5em}
    \gamma(\textbf{v}_n) = [sin(2\pi\textbf{B}\textbf{v}_n),cos(2\pi
    \textbf{B}\textbf{v}_n)]^{\top}
\vspace{-0.5em}
\end{split}
\vspace{-0.5em}
\]
The other is frequency tuning activations, represented by SIREN~\cite{sitzmann2020implicit}, using $sine(\omega\cdot x)$ instead of ReLU as $\phi$.

However, as stated above, both function expansion methods are limited in frequency representation thus their performance requiring further improvement. Moreover, MLP-based INRs~\cite{tancik2020fourier,sitzmann2020implicit,fathony2020multiplicative,saragadam2023wire} separately take each coordinate  $\textbf{v}_n$ into computation, ignoring their relationships and the locality prior.

\begin{figure*}[htbp]
  \centering
  \includegraphics[width=.99\textwidth]{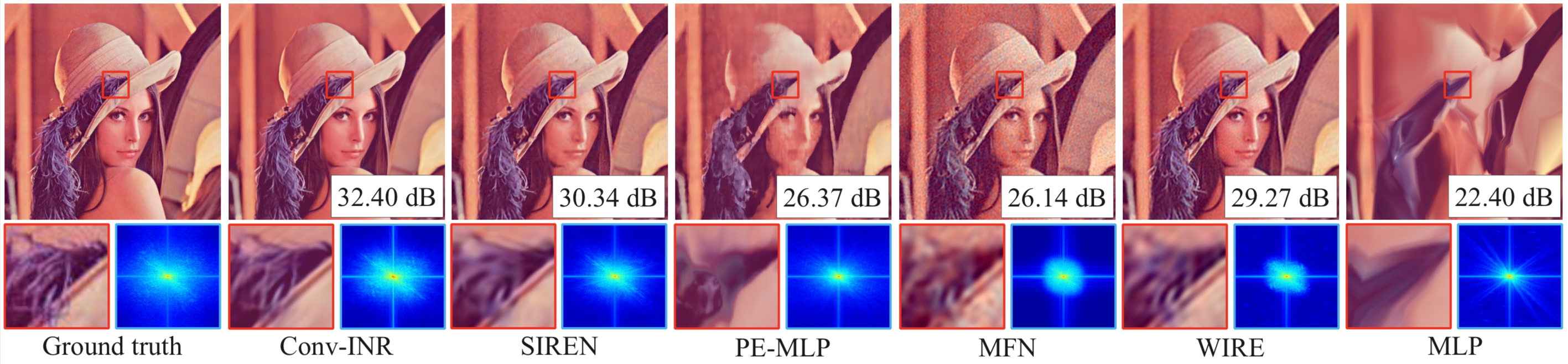} 
  \vspace{-1em}
  \caption{Comparisons of the Conv-INR and various MLP-based INRs for representing the 2D image Lena. The corresponding Fourier spectra are also visualized.}
  \label{p1}
  \vspace{-1em} 
\end{figure*}

\begin{figure*}[htbp]
  \centering
  \includegraphics[width=.99\textwidth]{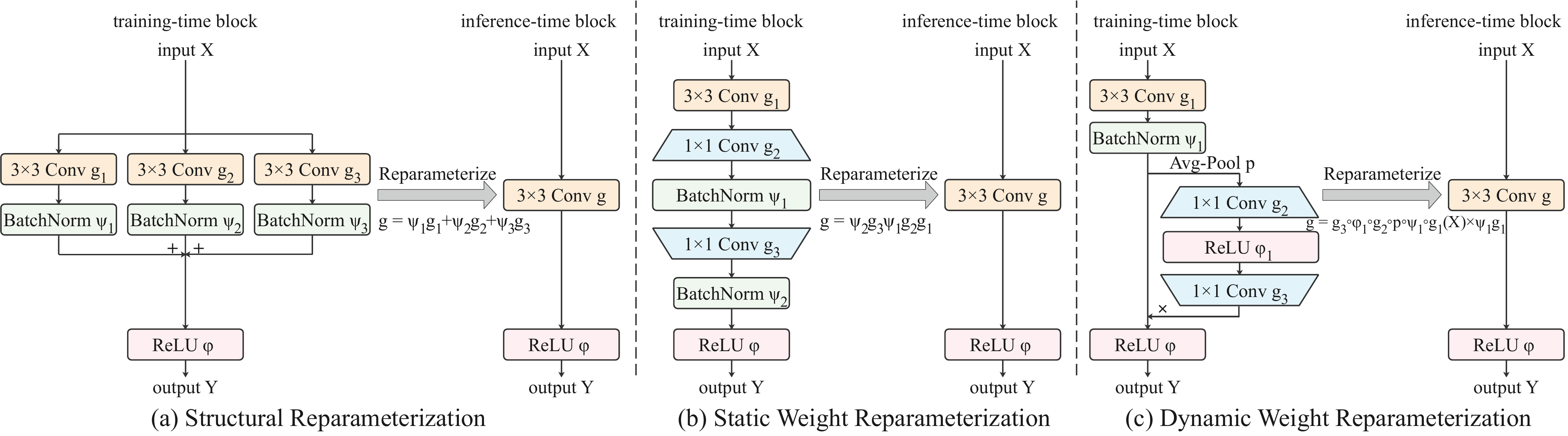} 
  \vspace{-1em}
  \caption{Reparameterization methods specifically tailored for Conv-INR. }
  \label{p2}
  \vspace{-1.5em} 
\end{figure*}

\vspace{-1em}
\section{Proposed Method}
\vspace{-0.5em}
\subsection{Convolutional INR}
\vspace{-0.5em}
To overcome the above issues, we raise Conv-INR especially for visual signals, which is the first INR method fully based on convolution and has better representational capacity. 
As shown in Fig~\ref{p0}, Conv-INR is alternatively consisting of $L$ convolution layers $g^l$, batch normalization layers $\psi^l$, and activation functions $\phi^l$.
For MLP-based INR, the total input is a $N\times C$ matrix $\textbf{V}$ stacked with $N$ coordinate vectors $\textbf{v}_n$ of length $C$ ($N\!=\!W\!\times\!H, C=2$ for image), and the total output is a $N\times C_{o}$ matrix $\textbf{I}(\textbf{V})$ stacked with $N$ coordinate vectors $I(\textbf{v}_n)$ of length $C_{o}$ ($C_{o}=3$ for image), finally, $\textbf{I}$ is reshaped to the same size as the target signal. 
For Conv-INR, $\textbf{V}$ is equivalently arranged to a 3D tensor $\Tilde{\textbf{V}}$ of $W\!\times\!H\!\times\!C$, which is the same spatial shape as the target visual signal, and it directly outputs the $W\!\times\!H\!\times\!C_o$ tensor $\Tilde{\textbf{I}}(\Tilde{\textbf{V}})$, which is exactly the represented signal. 
Thus the mathematical formulation of Conv-INR can be written as:
\vspace{-0.7em}
\[
\vspace{-0.5em}
    \begin{split}
\vspace{-0.5em}
    F^{conv}_{\theta}(\Tilde{\textbf{V}}) \!=\! g^L\!\circ\!\phi^{L\!-\!1}\!\circ\!\psi^{L\!-\!1}\!\circ\!g^{L\!-\!1}\!\circ\!\cdot\cdot\cdot\!\circ\!\phi^{1}\!\circ\!\psi^{1}\!\circ\! g^{1}(\Tilde{\textbf{V}})
    \label{eq3}
\vspace{-0.5em}
    \end{split}
\vspace{-0.5em}
\]
and the optimization target becomes:
\vspace{-0.7em}
\[
\vspace{-0.5em}
    \begin{split}
\vspace{-0.5em}
    \mathop{\arg\min}\limits_{\theta} \left|F^{conv}_{\theta}(\Tilde{\textbf{V}})-\Tilde{\textbf{I}}(\Tilde{\textbf{V}})\right|^2_2
    \label{eq4}
\vspace{-0.5em}
    \end{split}
\vspace{-0.5em}
\]

\textbf{Convolution leverages the locality prior.} 
Unlike MLP-based INRs using fully-connected layers, Conv-INR uses convolution layers as the core component, conducting convolution with a kernel size of $K\times K$ ($K=3$ by default) on the corresponding input, thus Conv-INR considers patches of adjacent pixels together instead of in isolation.
Note that the visual signals, such as images, usually exhibit strong local structures and tight neighborhood relationships.
Convolution can effectively model local dependency and leverage the locality prior, which is consistent with the inherent attributes of the visual signal, helping Conv-INR reconstruct high-quality signals with distinct local details as shown in Fig.~\ref{p1}.
As a matter of fact, MLP-based INR can be regarded as a special case of Conv-INR with a kernel size of $K=1$.

\textbf{Convolution overcomes the spectral bias.} 
Since convolution extracts local features (e.g., edges and textures), convolutional neural networks have an inherent capability to represent high-frequency signals~\cite{si2022inception,yin2019fourier}, which means Conv-INR is naturally free from the issue of spectral bias. 
Moreover, the positional encoding (PE) is theoretically equivalent to acting as a shift-invariant convolution kernel over the input domain, which is the reason for the effectiveness of PE~\cite{tancik2020fourier}. 
While Conv-INR directly leverages the inherent sliding window mechanism of convolution which conducts the shift-invariant operations on the input feature maps, thus addressing the issue of spectral bias in a more direct and thoroughgoing way than PE does. 
As shown in Fig.~\ref{p1}, Conv-INR can effectively represent wider range of spectral than other MLP-based INRs.
To supplement, Conv-INR does not require complex primary function expansions such as positional encoding or frequency tuning activations. Consequently, Conv-INR utilizes traditional ReLU as the nonlinear activation function.

\vspace{-1em}
\subsection{Reparameterization Techniques for Conv-INR}
\vspace{-0.5em}
Note that the vanilla Conv-INR has performed much better than existing MLP-based INRs as will be validated in Sec.~\ref{exps}. Nevertheless, we intend to further enhance the representational capacity of Conv-INR. Thus we introduce the reparameterization techniques to the realm of INR for the first time. The principle of reparameterization is to introduce more learnable parameters in training process, and then equivalently fuse these extra parameters into the original parameters after training, bringing enhanced representational capacity without incurring extra inference costs. We raise three reparameterization methods specially tailored for Conv-INR.

\textbf{Structural Reparameterization.} 
The core idea of structural reparameterization (SR) is to construct multiple convolution branches parallel to the main branch in training to extract more versatile representations, then due to the \textit{additivity} and \textit{homogeneity} of convolution, these extra branches can be equivalently fused into the main branch after training though directly adding their convolution kernels, thus improving the model representational capacity without extra inference cost~\cite{ding2019acnet,ding2021diverse}. 
To transfer SR to the Conv-INR, during the training time, we construct three parallel convolution layers $g_1,g_2,g_3$, each followed by a batch normalization layer $\psi_1,\psi_2,\psi_3$, then add the outputs of these three branches as shown in Fig.~\ref{p2}a. Given the input tensor $\textbf{X}$, the output of the block is $\textbf{Y}=\phi(\psi_1\!\circ\!g_1(\textbf{X})\!+\!\psi_2\!\circ\!g_2(\textbf{X})\!+\!\psi_3\!\circ\!g_3(\textbf{X}))$. 
After training, we first fuse the batch normalization layer into the corresponding convolution layer due to the \textit{homogeneity} (refer to \cite{ding2019acnet} for details) noted as $\psi_ig_i$, then we add the extra two convolution branches to the main convolution layer $g=\psi_1g_1+\psi_2g_2+\psi_3g_3$ due to the \textit{additivity}. We equivalently use $g$ for inference, and $\textbf{Y}=\phi\circ g(\textbf{X})$, thus maintaining the performance enhancement without introducing extra inference costs or altering the original model architecture.

\textbf{Static Weight Reparameterization.} 
We also introduce a specifically designed static weight reparameterization~\cite{cai2023refconv,arora2018optimization} for Conv-INR. 
As shown in Fig.~\ref{p2}b, we add two more $1\!\times\!1$ convolution layers $g_2,g_3$ (each followed with a batch normalization layer) right after the first $3\!\times\!3$ convolution layer $g_1$ with $C$ input/output channels. To enhance the representational capacity, these two $1\!\times\!1$ convolution layers follow the design of inverted bottleneck structure~\cite{sandler2018mobilenetv2}, i.e., the input channel of $g_2$ is $C$, its output channel is expanded to $4C$, and the output channel of $g_3$ is shrunk back to $C$. 
After training, we first individually fuse the batch normalization layers into $g_2,g_3$, and obtain two reparameterized $1\!\times\!1$ convolution layers $\psi_1g_2, \psi_2g_3$. Then according to the matrix multiplication~\cite{ding2021resrep}, we fuse $\psi_1g_2, \psi_2g_3$ into a new $1\!\times\!1$ convolution layer $\psi_2g_3\psi_1g_2$ with $C$ input/output channels, and finally fuse $\psi_2g_3\psi_1g_2$ to the first $3\!\times\!3$ convolution layer $g_1$, obtaining the reparameterized $3\!\times\!3$ convolution layer $g=\psi_2g_3\psi_1g_2g_1$ with $C$ input/output channels. Thus in inference, we use the equivalently transformed $g$ to conduct the convolution operation, namely, $\textbf{Y}=\phi\circ g(\textbf{X})$. 
 
\textbf{Dynamic Weight Reparameterization.} 
Typically, dynamic weight reparameterization (also regarded as attention mechanism) is data-dependent~\cite{cai2023refconv}, it dynamically generates different convolution kernels according to different inputs with hyper-networks~\cite{yang2019condconv,li2022omni,ha1609hypernetworks}. These hyper-networks can not be removed after training due to the constantly changing inputs. However, representing the signal with INR is an instance-aware task that each INR is trained specifically for a certain signal, thus the input is fixed when the training finished. Consequently, we design a dynamic weight reparameterization method specifically for Conv-INR that the hyper-network can be equivalently fused into the main convolution layer as shown in Fig.~\ref{p2}c. We construct an extra structure similar to the SE module~\cite{hu2018squeeze} except that we use an inverted bottleneck design to enlarge the parameter space. Given the input of the hyper-network $\psi_1\circ g_1(\textbf{X}) \in \mathbf{R}^{W\!\times\!H\!\times\!C}$, we first globally average-pool (noted as $p$) it, and input the obtained $1\!\times\!1\!\times\!C$ vector to the hyper-network and accordingly generate the $1\!\times\!1\!\times\!C$ coefficient vector, which is multiplied to $\psi_1\circ g_1(\textbf{X})$ along the channel dimension and obtain the final output, namely, $\textbf{Y}=\phi(g_3\circ\phi_1\circ g_2\circ p\circ\psi_1\circ g_1(\textbf{X}) \!\times\! \psi_1\circ g_1(\textbf{X}))$.
When the training finished, the input $\textbf{X}$ of each block (layer) is fixed, thus the coefficient vector of each layer is also fixed and can be numerically calculated. Consequently, we can discard the hyper-networks and directly multiple the coefficients to the corresponding channels of the batch norm fused convolution kernel $\psi_1 g_1$. Thus we can obtain the final reparameterized convolution kernel $g=g_3\circ\phi_1\circ g_2\circ p\circ\psi_1\circ g_1(\textbf{X}) \!\times\! \psi_1 g_1$.

\begin{figure*}[!t]
  \centering
  \begin{subfigure}{0.24\linewidth}
    \includegraphics[width=0.98\linewidth]{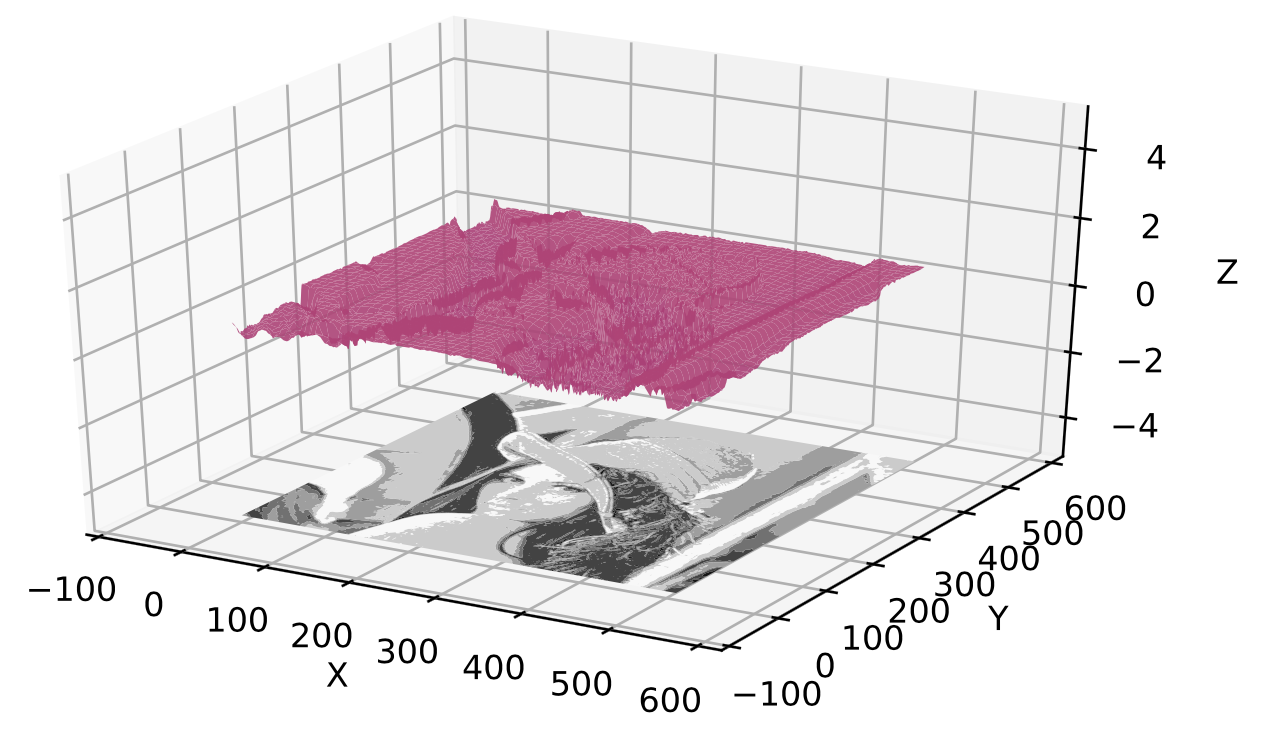}
    \caption{Target Image Lena}
    \label{p2-1}
  \end{subfigure}
  \begin{subfigure}{0.24\linewidth}
    \includegraphics[width=0.98\linewidth]{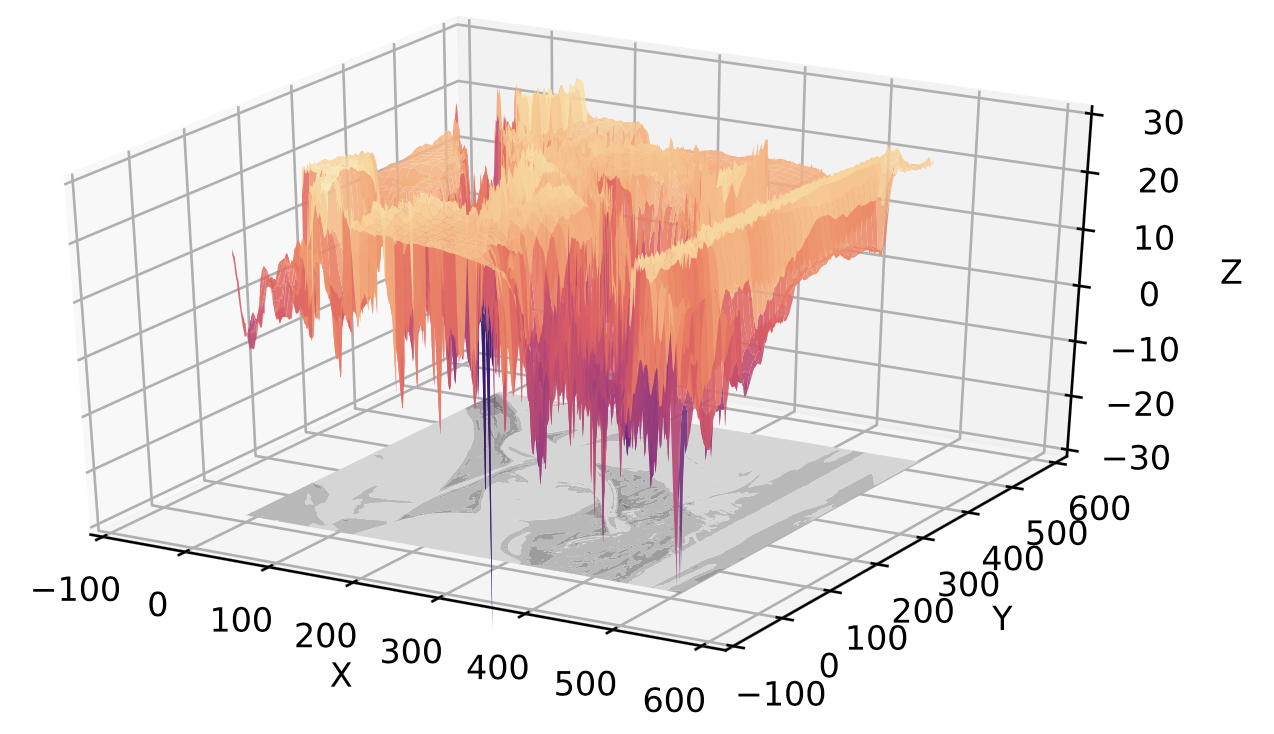}
    \caption{Conv-INR Linear}
    \label{p2-2}
  \end{subfigure}
    \begin{subfigure}{0.24\linewidth}
    \includegraphics[width=0.98\linewidth]{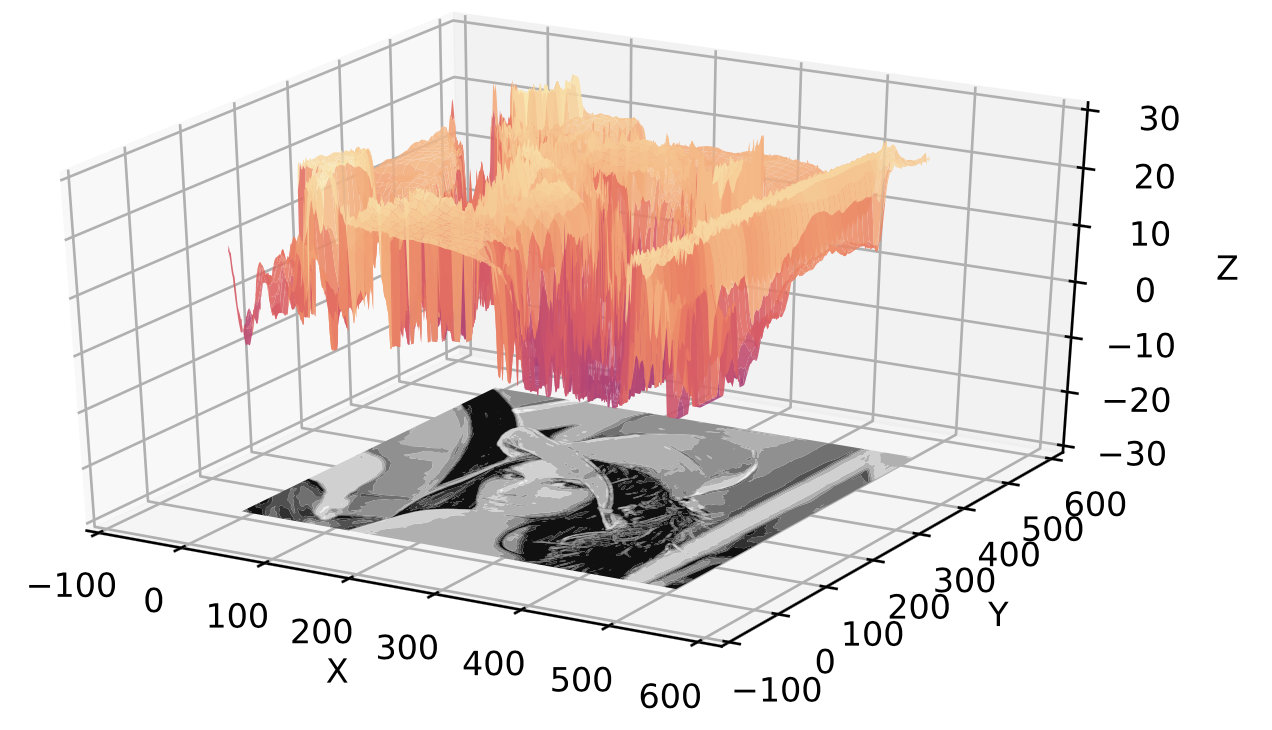}
    \caption{Conv-INR Activation}
    \label{p2-3}
  \end{subfigure}
  \begin{subfigure}{0.24\linewidth}
    \includegraphics[width=0.98\linewidth]{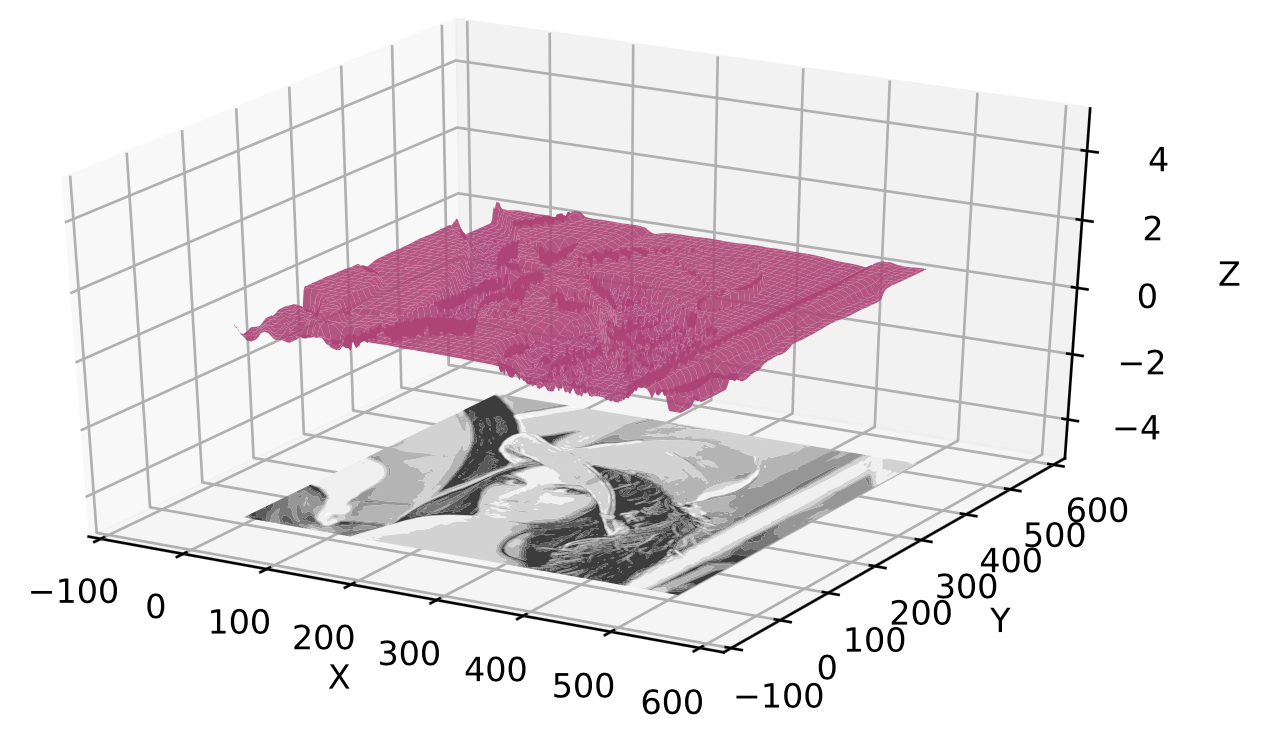}
    \caption{Conv-INR Normalization}
    \label{p2-4}
  \end{subfigure}
  \begin{subfigure}{0.24\linewidth}
    \includegraphics[width=0.98\linewidth]{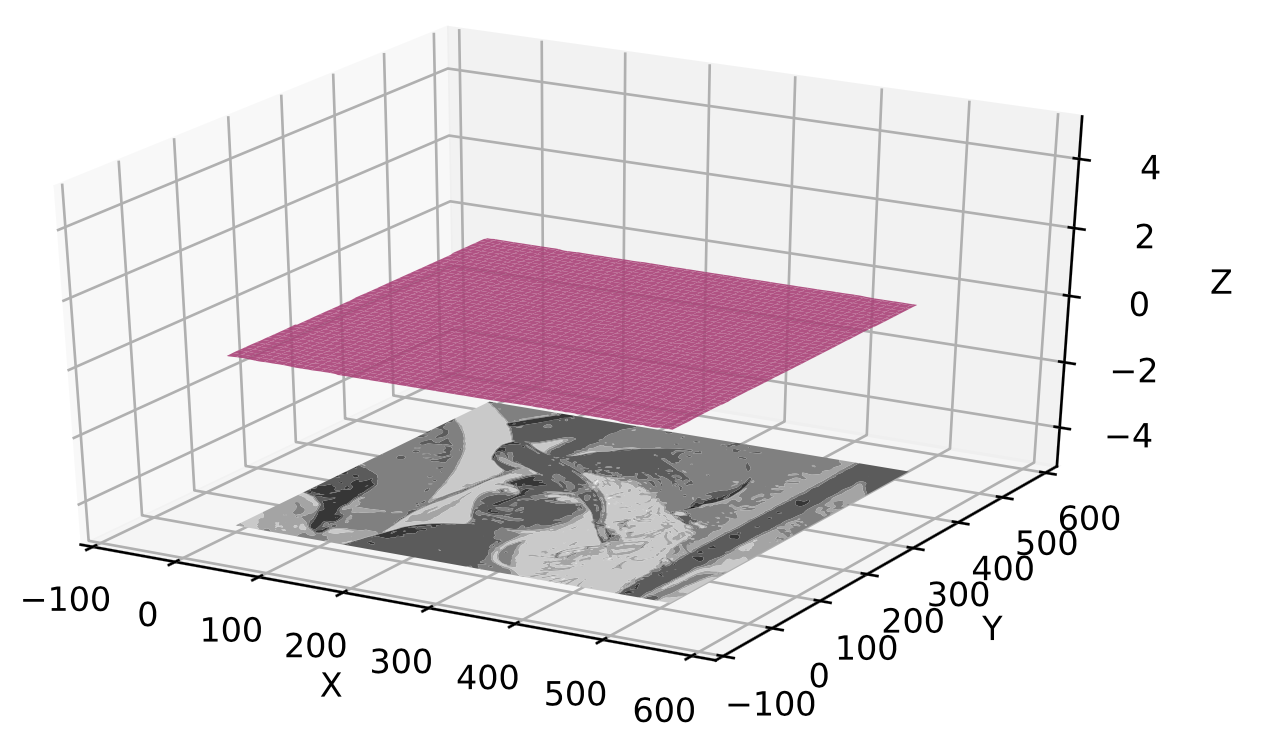}
    \caption{SIREN Linear}
    \label{p2-5}
  \end{subfigure}
  \begin{subfigure}{0.24\linewidth}
    \includegraphics[width=0.98\linewidth]{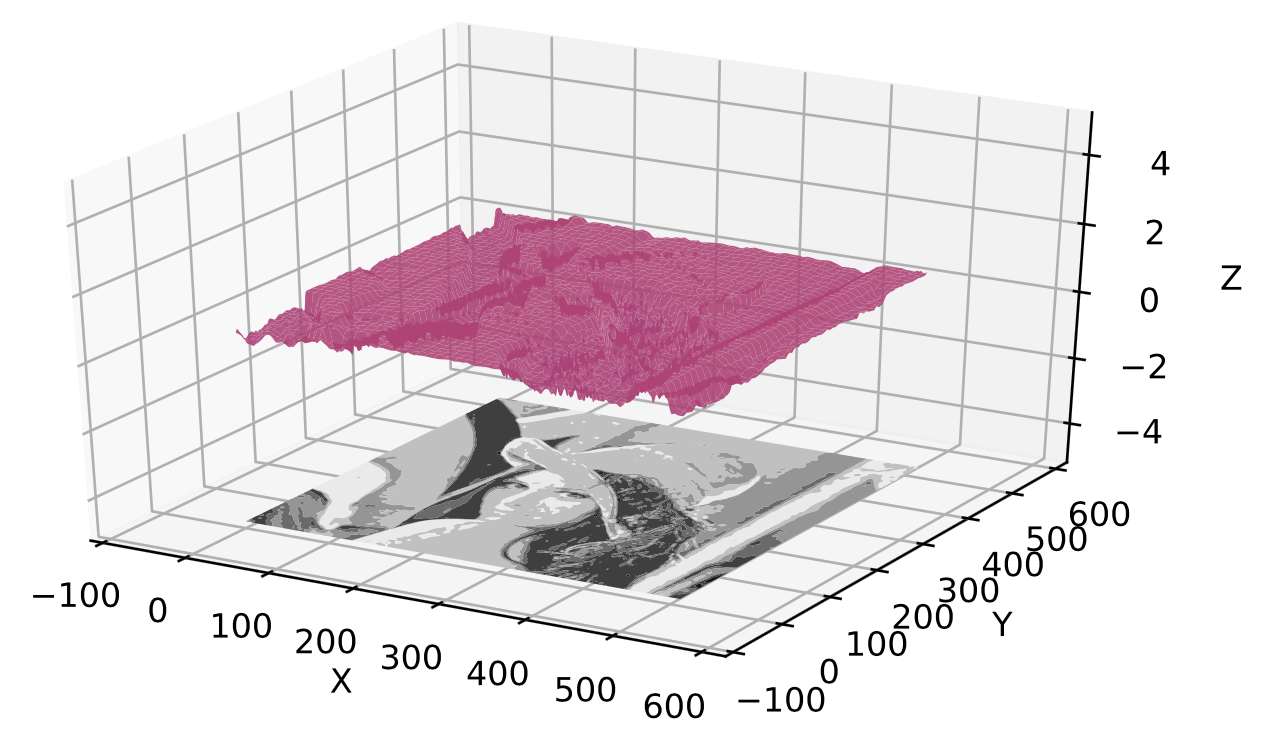}
    \caption{SIREN Activation}
    \label{p2-6}
  \end{subfigure}
  \begin{subfigure}{0.24\linewidth}
    \includegraphics[width=0.98\linewidth]{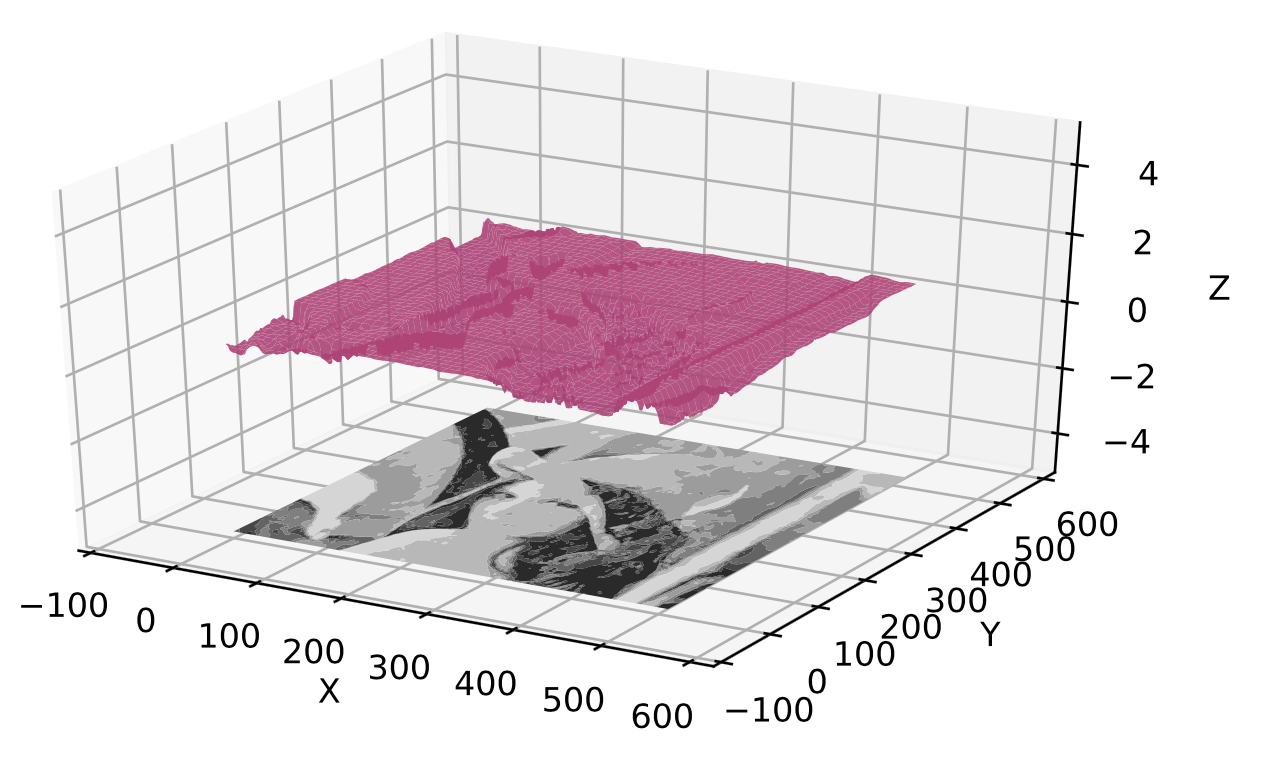}
    \caption{PE-MLP Linear}
    \label{p2-7}
  \end{subfigure}
  \begin{subfigure}{0.24\linewidth}
    \includegraphics[width=0.98\linewidth]{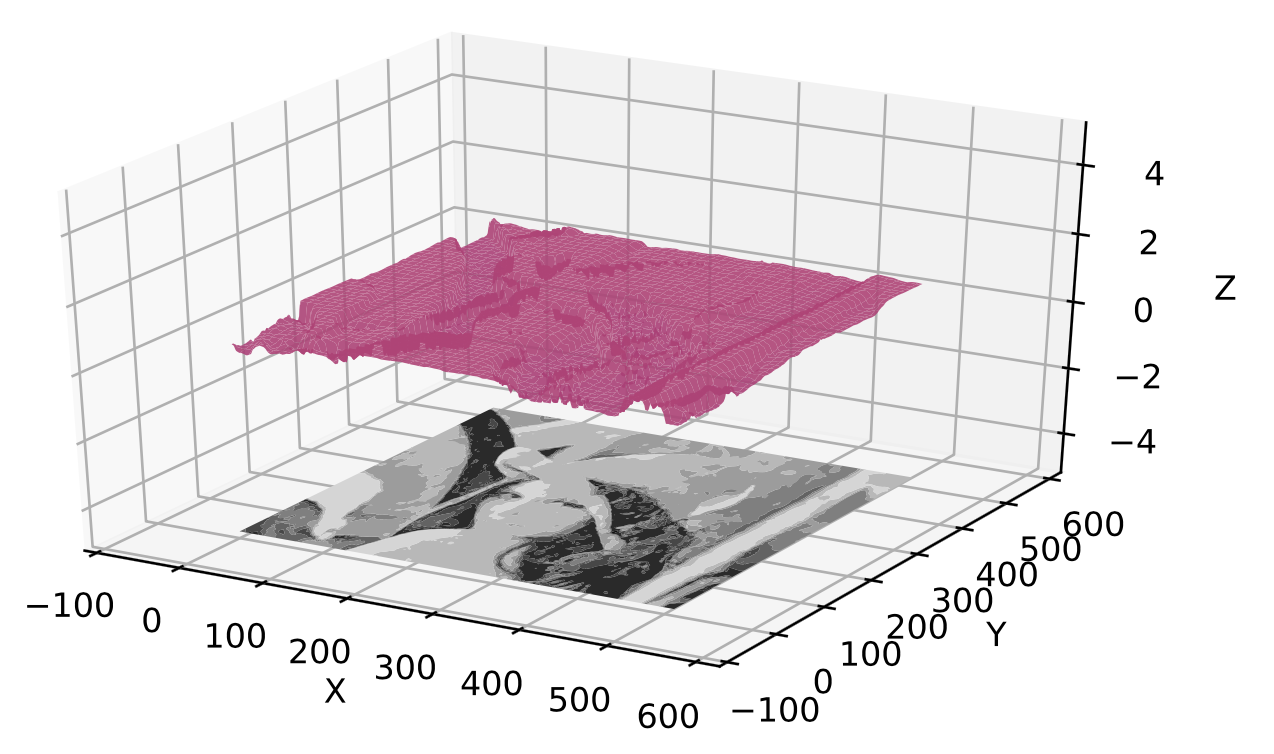}
    \caption{PE-MLP Activation}
    \label{p2-8}
  \end{subfigure}
  \vspace{-1em} 
  \caption{3D intensity of the feature maps obtained by Conv-INR, SIREN and PE-MLP.}
  \label{p3}
  \vspace{-1em}
\end{figure*}

\begin{table}[t]
\centering
\begin{tabular}{l|cccc}
\hline
Model        & 2D Image & 2D CT & 3D MRI & 5D Nerf \\
\hline
MLP          & 21.12    & 26.31 & 24.52  & 15.12 \\
SIREN        & 25.52    & 28.30 & 26.04  & 25.44 \\
PE-MLP       & 23.16    & 28.11 & 30.17  & 30.99 \\
MFN          & 25.25    & 27.97 & 27.24  & 31.04 \\
WIRE         & 25.05    & 28.26 & 25.31  & 25.76 \\
\textbf{Conv-INR }    & \textbf{27.01}    & \textbf{29.72} & \textbf{31.64}  & \textbf{31.78} \\
\hline
\textbf{SR Conv-INR}  & 27.25    & 29.83 & 31.84  & 32.09 \\
\textbf{WR Conv-INR}  & 27.67    & 29.97 & 31.96  & 31.95 \\
\textbf{DR Conv-INR}  & 27.41    & 29.93 & 32.02  & 32.07 \\
\hline
\end{tabular}
\vspace{-0.5em} 
\caption{Results of vanilla Conv-INR, MLP-based INRs and Conv-INR with reparameterization methods on four tasks.}
\label{t1}
\vspace{-1.5em}
\end{table}

\vspace{-1em}
\section{Experiments}
\label{exps}
\vspace{-0.5em}
We validate the effectiveness of Conv-INR on four separate tasks, \textit{i.e.}, 2D image fitting and compression, 2D computed tomography (CT) reconstruction, 3D magnetic resonance imaging (MRI) reconstruction, and 5D novel view synthesis.
The kernel size of Conv-INR is set as $K=3$ by default.
We compare Conv-INR with the vanilla MLP INR and four popular MLP-based INRs, namely, PE-MLP~\cite{tancik2020fourier},  SIREN~\cite{sitzmann2020implicit}, MFN~\cite{fathony2020multiplicative} and WIRE~\cite{saragadam2023wire}.
As a common practice, the encoding scale of PE is set as $10$~\cite{mildenhall2021nerf}, the frequency parameter $\omega_0$  of SIREN is set as $30$~\cite{sitzmann2020implicit}, the frequency parameter $\omega$ and the spread parameter $s$ of WIRE are respectively set as $20$ and $10$~\cite{saragadam2023wire}.
The weights of all the INRs are randomly initialized. For SIREN and MFN, we utilize the specific weight initialization schemes as raised in \cite{sitzmann2020implicit} and \cite{fathony2020multiplicative}, respectively. For the left INR methods, we utilize the default LeCun random initialization~\cite{lecun2002efficient}.

\vspace{-1.5em}
\subsection{Image Fitting}
\vspace{-0.5em}
We first use an 2D image fitting task to evaluate the performance of Conv-INR. 
We perform experiments on the Kodak dataset~\cite{dupont2021coin} consisting of 24 RGB images with a high resolution of $768\times512$.
We use the networks with $10$ hidden layers and each layer has $32$ channels. 
All the models are trained for 100,000 iterations using Adam optimizer~\cite{kingma2014adam} with an initial learning rate of $2e\!-\!4$.
Table.~\ref{t1} shows the average experimental results of these INRs measured in PSNR (Peak Signal-to-Noise Ratio). As can be observed, Conv-INR significantly surpasses existing MLP-based INR methods. Specifically, Conv-INR exceeds the vanilla MLP INR up to $5.89$ dB.
Moreover, Conv-INR surpasses PE-MLP and SIREN by a clear margin of $3.85$ and $1.49$ dB, respectively. 

\textbf{Visualization of the reconstructed images.}
We visualize the reconstructed images and corresponding Fourier frequency spectra of various methods in Fig.~\ref{p1}. The target visual signal is 2D image Lena with a resolution of $512\times512$.
As can be obviously observed, the reconstructed image of ReLU is over-smoothed and poor-quality, failing to present the details of the target image. Simultaneously, ReLU can only limitedly represent low frequencies, highlighting the presence of spectral bias. 
For PE, the quality of the reconstructed image is poor for that the overall image is blurry and the finer details are failed to be restored. Moreover, both low- and high-frequencies are learned limitedly.
While for Conv-INR, the quality of the reconstructed images is significantly improved, and the details are clearly represented. 
Furthermore, as clearly shown by the Fourier spectra, more low- and high-frequency components are learned effectively, illustrating the effectiveness of Conv-INR on alleviating the spectral bias.

\textbf{Visualization of the 3D Intensity.}
We further visualize the 3D intensity of the feature maps obtained by the final linear layers and activation layers of Conv-INR, SIREN and PE-MLP, and the final normalization layer of Conv-INR. 
We can observe that the linear (convolution) layer of Conv-INR can learn abundant high-frequency signals since the 3D intensity fluctuates rapidly (Fig.~\ref{p2-2}), while the linear layer of SIREN fail to learn any high-frequency signals since the 3D intensity is extremely flat (Fig.~\ref{p2-5}). The normalization layer in Conv-INR then flattens the rapid fluctuated intensity into target range (Fig.~\ref{p2-4}). While the activation layer in SIREN adds fluctuations to the intensity which introduces different frequencies (Fig.~\ref{p2-6}). 

\vspace{-1.5em}
\subsection{CT Reconstruction.}
\vspace{-0.5em}
In 2D CT task, we train an INR that takes in 2D pixel coordinates and predicts the corresponding volume density at each location.
We conduct the experiments on the x-ray colorectal dataset~\cite{saragadam2023wire,clark2013cancer}, each image has a resolution of $512\times 512$ and is emulated with 100 CT measurements.
We use networks with 2 hidden layers and 256 channels.
All the models are trained for 20,000 iterations using Adam optimizer with a initial learning rate of $5e\!-\!3$.
Table.~\ref{t1} provides the experimental results measured in PSNR. As can be observed, Conv-INR  consistently achieves the highest PSNR among all the INRs. 

\vspace{-1.5em}
\subsection{MRI Reconstruction.}
\vspace{-0.5em}
For the 3D MRI task, we train an INR that takes in 3D voxel coordinates and predicts the corresponding intensity at each location. 
We conduct experiments on the ATLAS brain dataset~\cite{tancik2020fourier}, each sample has a volume resolution of $96^3$.
We use networks with 2 hidden layers and 256 channels. 
All the models are trained for 1,000 iterations using Adam optimizer with an initial learning rate $2e\!-\!3$. 
As shown in Table.~\ref{t1}, Conv-INR consistently achieves better performance than other MLP-based INRs.  

\vspace{-1em}
\subsection{Novel View Synthesis.}
\vspace{-0.5em}
For the novel view synthesis using 5D neural radiance fields (NeRF)~\cite{mildenhall2021nerf} experiments, the input contains the 3D position and 2D viewing direction of a point and the output attributes include the RGB color and point density.
We only use Conv-INR for the prediction of the RGB color.
We follow the model architecture and training configuration in \cite{mildenhall2021nerf}, and conduct the experiments on the NeRF dataset~\cite{mildenhall2021nerf}. 
As can be observed in Table.~\ref{t1}, Conv-INR  consistently achieves better performance than other MLP-based INRs.

\vspace{-1em}
\subsection{Reparameterizations for Performance Enhancement}
\vspace{-0.5em}
We further conduct extensive experiments to validate the effectiveness of the three reparameterization methods specially raised for Conv-INR, namely, structural reparameterization (SR), static weight reparameterization (WR), and dynamic weight reparameterization (DR).
We use LeCun initialization to initialize the additionally introduced weights and follow the same training configurations as stated above on these four tasks.
Table.~\ref{t1} exhibits the experimental results. 
To emphasize that, these additionally introduced structures and parameters can be equivalently transformed into a single reparameterized convolution layer with the same original shape when training finished as stated above, thus improving the performance with no extra parameters or computations in inference.

\vspace{-1em}
\section{Conclusion}
\vspace{-0.5em}
This paper propose the first INR backbone model based on convolution, termed as Conv-INR, especially for visual signals.
Conv-INR adheres to the intrinsic attributes of visual signals and naturally overcomes the issue of frequency-bias.
It has better representational capacity and trainability than existing MLP-based INRs.
We also propose three reparameterization methods specifically tailored for Conv-INR, which further enhance the representational capacity without any inference cost. 
Extensive experiments on four challenging tasks validate the effectiveness of Conv-INR.

{
\small


}
\end{document}